\documentclass[a4paper,12pt]{article}
\usepackage{mathptmx}
\usepackage{amsmath,amstext,amsgen,amsbsy,amsopn,amsfonts,amssymb}
  \usepackage{setspace}
\usepackage{easybmat}
\usepackage{graphics}
\usepackage[pdftex]{graphicx}
\usepackage{epsfig}
\usepackage{amsthm}
\usepackage{bm}
\usepackage{algorithm}
\usepackage{algorithmic}

\usepackage{wrapfig}
\usepackage{esvect}
\usepackage{multirow}
\usepackage{array}
\usepackage{xcolor}
\usepackage{hyperref}

\begin{document}
\large
\title{Early Prediction of Geomagnetic Storms by \\Machine Learning Algorithms}
\author{
Iris Yan
\vspace{0.1in}\\
Amity Regional High School\vspace{0.06in}\\
Woodbridge, CT 06525\\
USA
}
\date{}
\maketitle
\normalsize
\tableofcontents
\newpage

\begin{abstract}
\noindent
Geomagnetic storms (GS) occur when solar winds disrupt Earth’s magnetosphere. GS can cause severe damages to satellites, 
power grids, and communication infrastructures. Estimate of direct economic impacts of a large scale GS exceeds \$40 billion a 
day in the US. Early prediction is critical in preventing and minimizing the hazards. However, current methods either predict several 
hours ahead but fail to identify all types of GS, or make predictions within short time, e.g., one hour ahead of the occurrence. This work aims to 
predict all types of geomagnetic storms reliably and as early as possible using big data and machine learning algorithms. By fusing big data collected from 
multiple ground stations in the world on different aspects of solar measurements and using Random Forests regression with feature selection and downsampling 
on minor geomagnetic storm instances (which carry majority of the data), we are able to achieve an accuracy of  82.55\% on data collected 
in 2021 when making early predictions three hours in advance. Given that important predictive features such as historic Kp indices are measured every 3 hours 
and their importance decay quickly with the amount of time in advance, an early prediction of 3 hours ahead of time is believed to be close to the practical limit.
\end{abstract}

\section{Introduction}
\label{section:introduction}
\begin{figure}[!htb]
\centering
\begin{center}
\includegraphics[scale=0.5,clip, angle=0]{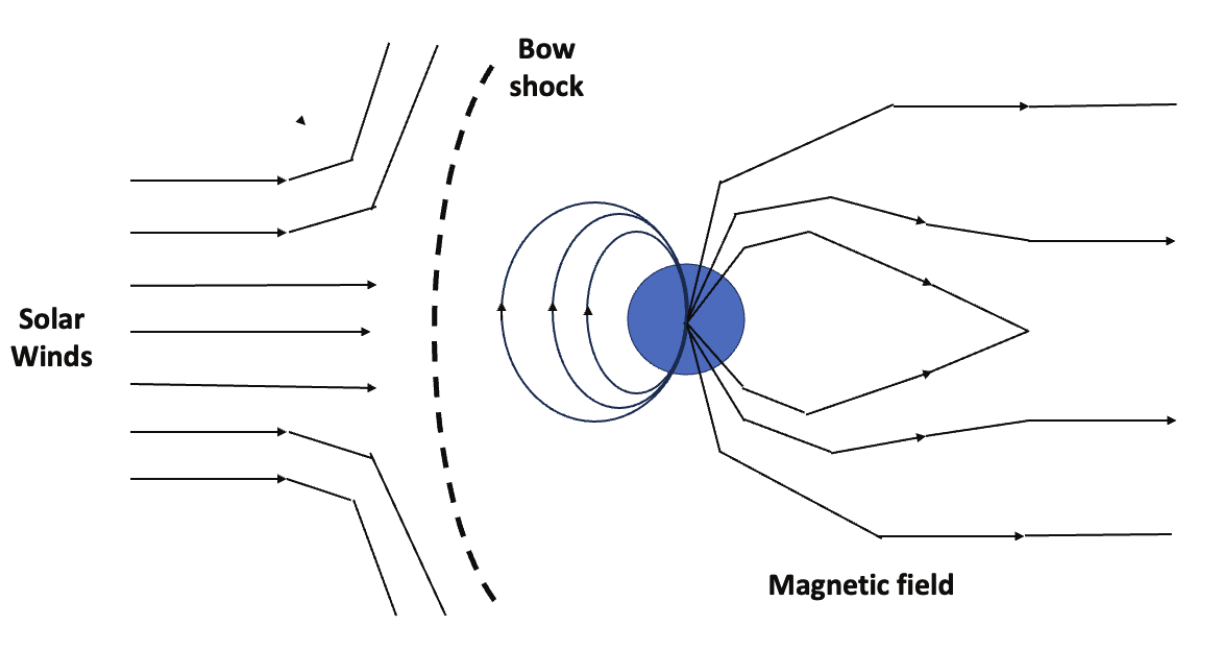}
\end{center}
\abovecaptionskip=-1pt
\caption{\it Illustration of solar winds, Earth's magnetosphere and geomagnetic storms. The dark disc indicates the Earth. 
The dashed line indicates the surface of the Earth's magnetosphere, where bow shock 
occurs when the solar winds interact with the Earth's magnetosphere.} 
\label{figure:gms}
\end{figure}

\begin{figure}[h]
\centering
\begin{center}
\hspace{0cm}
\includegraphics[scale=0.33,clip,angle=0]{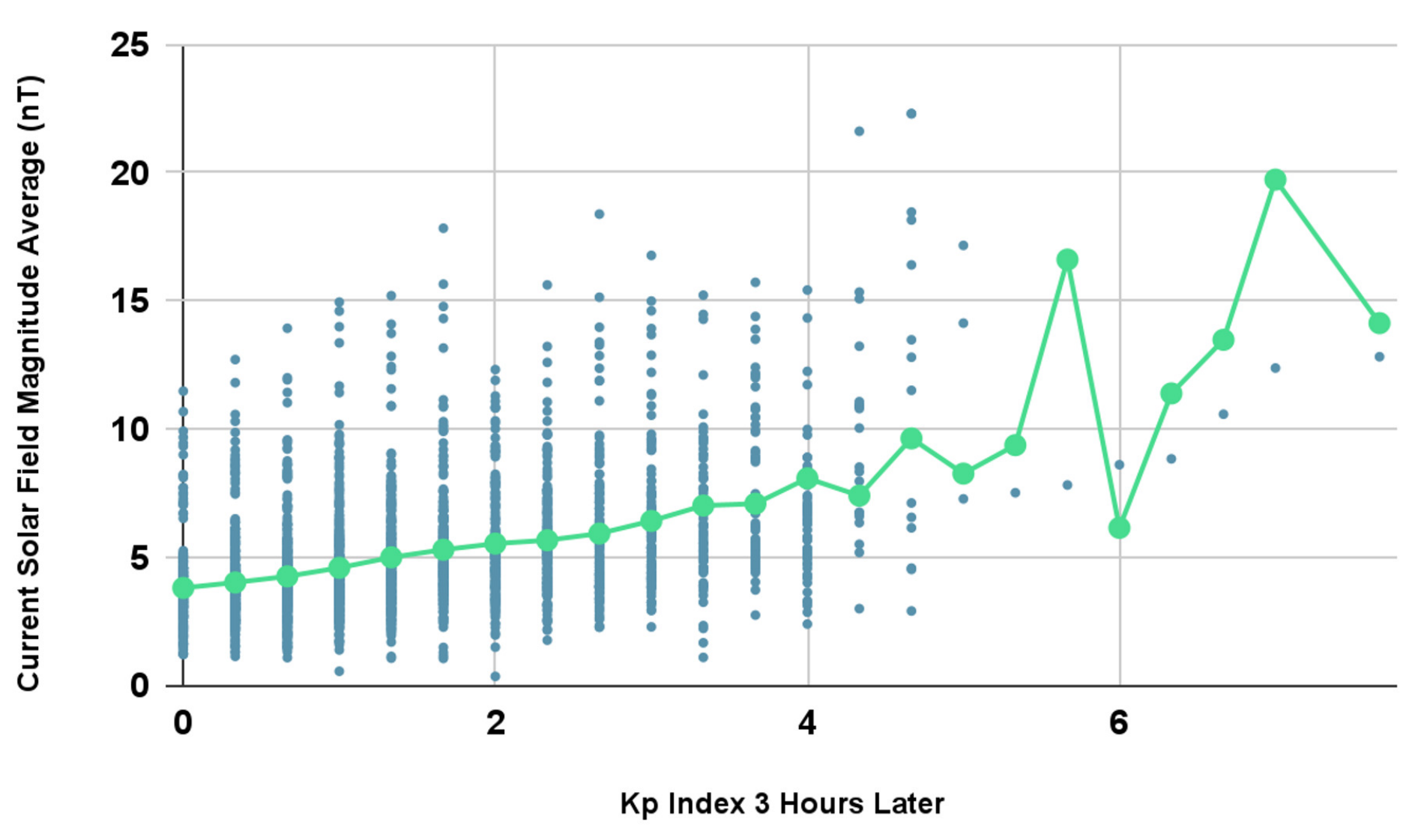}
\end{center}
\abovecaptionskip=-1pt
\caption{\it Wide range of solar field magnitudes for a given future (e.g., 3 hours later) Kp-index. } 
\label{figure:kpIndex}
\end{figure}

Geomagnetic storms (GS) are major disturbances to the Earth’s magnetosphere caused by solar winds \cite{NOAASwpcGS}. In Earth's atmosphere, there is a surrounding layer formed by Earth’s magnetic field called magnetosphere. 
The solar winds are created by high-temperature outgoing streams of plasma (i.e., charged particles) emitted from the solar flares, which continuously buffet and compress the Earth's magnetosphere. Geomagnetic storms occur when there is a sudden change of plasma movements, electric currents and hence magnetic forces in the Earth's magnetosphere. It is interesting to note that geomagnetic storms can also produce the beautiful aurora borealis one may observe in the sky. An illustration of the solar winds, Earth's magnetosphere and GS can be seen in Figure~\ref{figure:gms}. 
 \\
 \\
 During the occurrence of geomagnetic storms, there is a buildup and discharge of electrons which resembles lightning. GS can cause severe damages to satellites, 
 power grids, and communication infrastructures. Estimate of direct economic impacts of a large scale GS exceeds \$40 billion a 
 day in the US alone \cite{Oughton2017}. GS can cause electrical blackouts and internet outages on massive scales that may stay unrepaired for months. 
A noted recent occurrence of GS was on February 3--4, 2022, and it was reported that 38 out of 49 Starlink satellites were destroyed by an intense geomagnetic storm. 
Early prediction 
 is critical in preventing and minimizing the potential hazards that would be caused by GS. However, current methods either predict GS several hours ahead but fail to identify 
 all the major types \cite{Podladchikova2012}, or make predictions not early enough (i.e., only within one hour of the occurrence) \cite{ZhuBillings2006}.
This work aims to predict all types of geomagnetic storms reliably and as early as possible using machine learning algorithms.
\\
\\
Early predictions of geomagnetic storms is very challenging. It is nearly impossible to approach the problem from the physics of geomagnetic storms, due to its high complexity \cite{GonzalezJoselyn1994}. One has to use current (or historic) solar measurements to predict the 
geomagnetic disturbance in the future. However, current measurements of solar activity entail very little information about the magnitude
of future geomagnetic storms \cite{ElliottJahn2013}.
The intensity of geomagnetic storms is commonly measured by the Kp index which takes value in the range of 0 -- 9. For instance, the largest recorded geomagnetic storm, which occurred in 1859, had an estimated Kp index in the range of 8.2 -- 8.7 by Wagner and his coauthors \cite{Wagner2023}. The Feb 3-4, 2022 geomagnetic storm had an estimated Kp index around 5.0 -- 5.3 \cite{BojilovaMukharov2023}.
Figure~\ref{figure:kpIndex} shows that a given level of future (e.g., 3 hours later) Kp index can correspond to a wide range of values 
for current solar measurements.
\\
\\
Another challenge lies in the extreme sparsity of the signal and the highly unbalanced nature of the data in the sense that 
major storms (Kp indices $> 4$) 
occur rarely \cite{ChapmanHorne2019}. A quick inspection of historic data shows that major storms carry less than 3\%
of the data. To detect such extremely sparse signals is like finding a needle in haystacks. This can lead to the masking 
phenomenon in machine learning, where classes with a few data instances may be ignored by algorithm.
Additionally, as the data eventually collected at the ground stations are transmitted from satellites (which have sensors on board to 
collect measurements about various solar activities, geomagnetic disturbances etc), factors such as unfavorable 
weather conditions or interferences from electromagnetic signals can introduce substantial noises and 
distortion to data collected at the ground stations.
\\
\\
This work aims to overcome the above challenges by learning highly complex patterns using Random Forests (RF) regression \cite{RF}, 
and using techniques including downsampling and feature selection to further overcome technical challenges present
in the data. In the remaining of this paper, we describe our approach and implementation in Section~\ref{section:method}. This is 
followed by a report of experimental results in Section~\ref{section:experiments}. Finally we 
conclude in Section~\ref{section:conclusions}. 
%
\section{The method}
\label{section:method}
\begin{figure}[!htb]
\centering
\begin{center}
\includegraphics[scale=0.43,clip, angle=0]{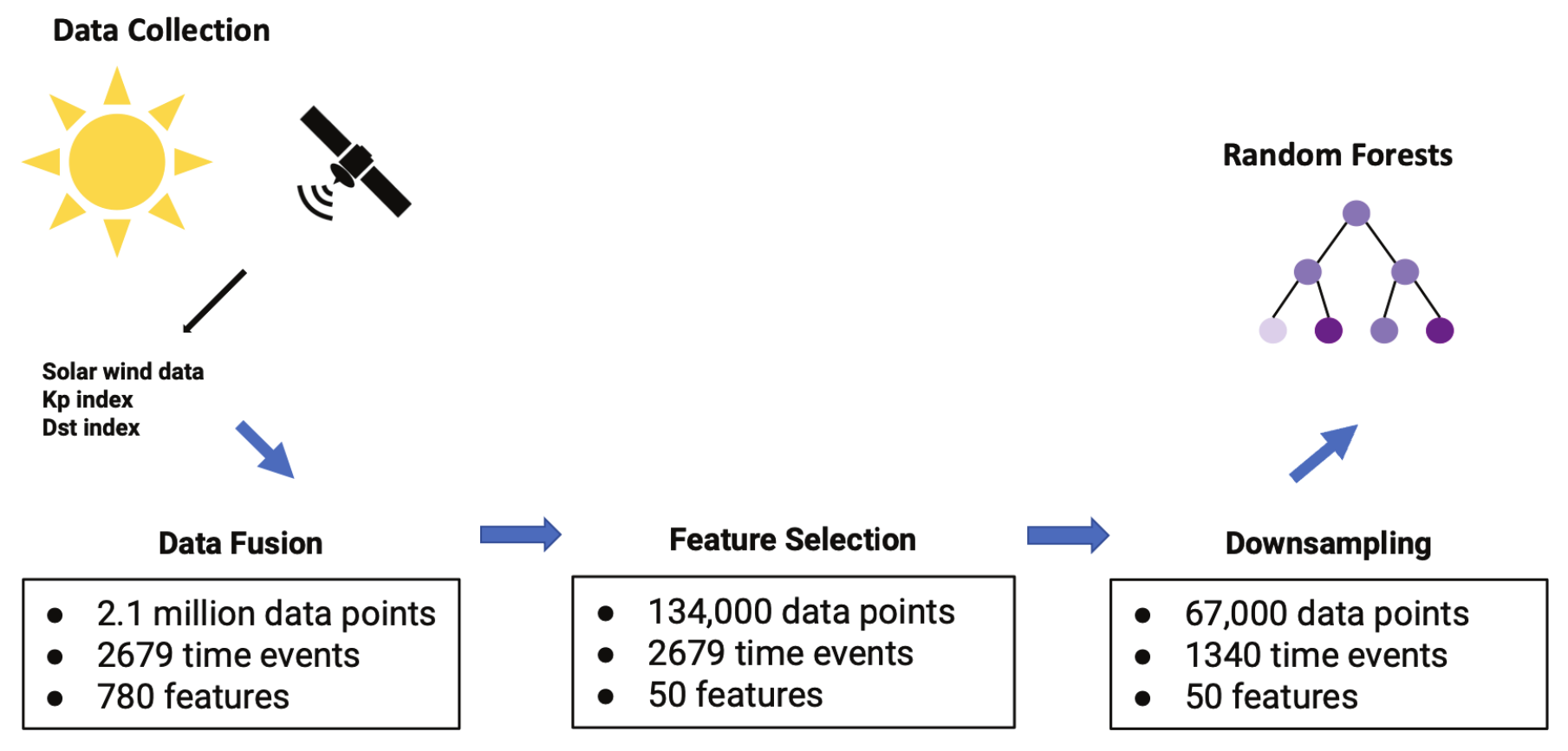}
\end{center}
\caption{\it Overall architecture of the proposed approach.} 
\label{figure:arch}
\end{figure}

The early prediction of geomagnetic storms is formulated as a prediction of future (i.e., 3 hours after the time of prediction) Kp index of the storm using past measurements of solar activities, past density and Kp indices. 
Our approach consists of data collection, data fusion, feature selection, downsampling and the training of Random Forests regression
model for prediction. Here, instead of trying very hard on improving the algorithm, we adopt techniques like feature selection and downsampling to ``improve" the datasets for the purpose of sparse pattern detection. Figure~\ref{figure:arch} is a illustration of our approach. For the rest of this section, we will describe individual steps in our approach. 
\\
\\
One important feature of this work is the use of big data collected from multiple sources in the analysis. We have measurements of
solar activities or intensities of geomagnetic disturbances from 
NASA OMNIWeb \cite{NOAASwpcGS, NOAASwpcSW, NASAspdf}, the Kyoto World Data Center in Tokyo, Japan \cite{WDCKyoto}, 
the German Center for Geosciences \cite{GFZGerman}, respecively. The OMNIWeb data includes solar wind speed, proton density, wind temperature, field magnitude averages, measured every 5 minutes, for up to 9 hours before the time of the prediction. The Kyoto data included past Dst (i.e., density) indices, measured every hour, for up to 3 hours before the time of the prediction. The GFZ data included past Kp indices, measured every 3 hours, up to 24 hours before the time of the prediction. 
\\
\\
Together the data translates to 2.1 million data points, 2679 time events and 780 features (variables). For each point,
there are 780 features associated with it, including {\it Kp indices, field magnitude average, solar wind direction vector components Bx, By, Bz, solar wind speed, 
electron and particle density, and solar wind temperature} for time intervals with feature-specific interval lengths during the last 12 hours till the current time, and the response variable is the Kp index three hours later from the time of prediction. Note that here the feature-specific time interval lengths are 5 minutes for features generated from the OMNIWeb data, one hour for those from the Kyoto data, and three hours for past Kp indices from the GFZ data.
\\
\\
The next step is feature selection. As there are 780 features after the initial 
data fusion, and some features may be more important than others and many features may not be as informative (or otherwise would
introduce noises) so
it might be better to discard them. This would help reduce potential wrong generalizations deduced from data. So we selectively include 
important features (out of 780) using the built-in feature selection tool provided by RF. RF produces a rank of individual features according to their importance, and we use the top 50 or 100 features according to this rank. A further step is downsampling to discard data from larger classes (so that the minority classes or those data points
associated with major storms carry a higher percentage of the data). This creates a more balanced distribution of data 
 across different classes, which can often improve accuracy in prediction.
\subsection{Random Forests regression}
\label{section:vignetteRF}
Random Forests \cite{RF} regression is the algorithm used for model fitting in our analysis. RF is an ensemble of decision trees with 
each tree constructed on a bootstrap sample of the data. Each tree is built by recursively partitioning the data. At each 
tree node (the root node corresponds to the bootstrap sample), RF randomly samples a number of features and then select 
one that would lead to an ``optimal" partition of that node. Each tree recursively narrows down which Kp index an instance 
should fall into. This process is continued recursively until the tree is fully grown. There are two modes in RF, classification
and regression, and we use regression in this work. That is, we grow each tree until each leaf node has at most 5 data points
or consists of data points of the same value. The fitted value for a data point falling into a node is taken as the average of values 
of all points in the same leaf node. The final decision of a forest is taken from the average of decisions from individual trees.
The superior empirical performance of RF has been demonstrated by a 
number of studies \cite{RF,Caruana2006, caruanaKY2008}. RF is easy to use 
(e.g., very few tuning parameters), has a remarkable built-in ability for feature selection. We use the R package  
{\it randomForest} in this work.
\\
\\
For prediction, we try to learn the relationship between the future Kp index and the current (or historic) solar measurements. 
This is a regression problem, and a simple choice would be to use linear regression. Will linear regression be a sensible choice 
to model the
relationship of interest? We conduct an exploratory analysis of the data by visualizing the data---to see how the data looks like.
As the raw data has a dimension of 780, we reduce the data dimension by principle component analysis and visualize the 
data by its top two principal directions. 
\begin{figure}[h]
\centering
\begin{center}
\hspace{0cm}
\includegraphics[scale=0.62,clip,angle=-90]{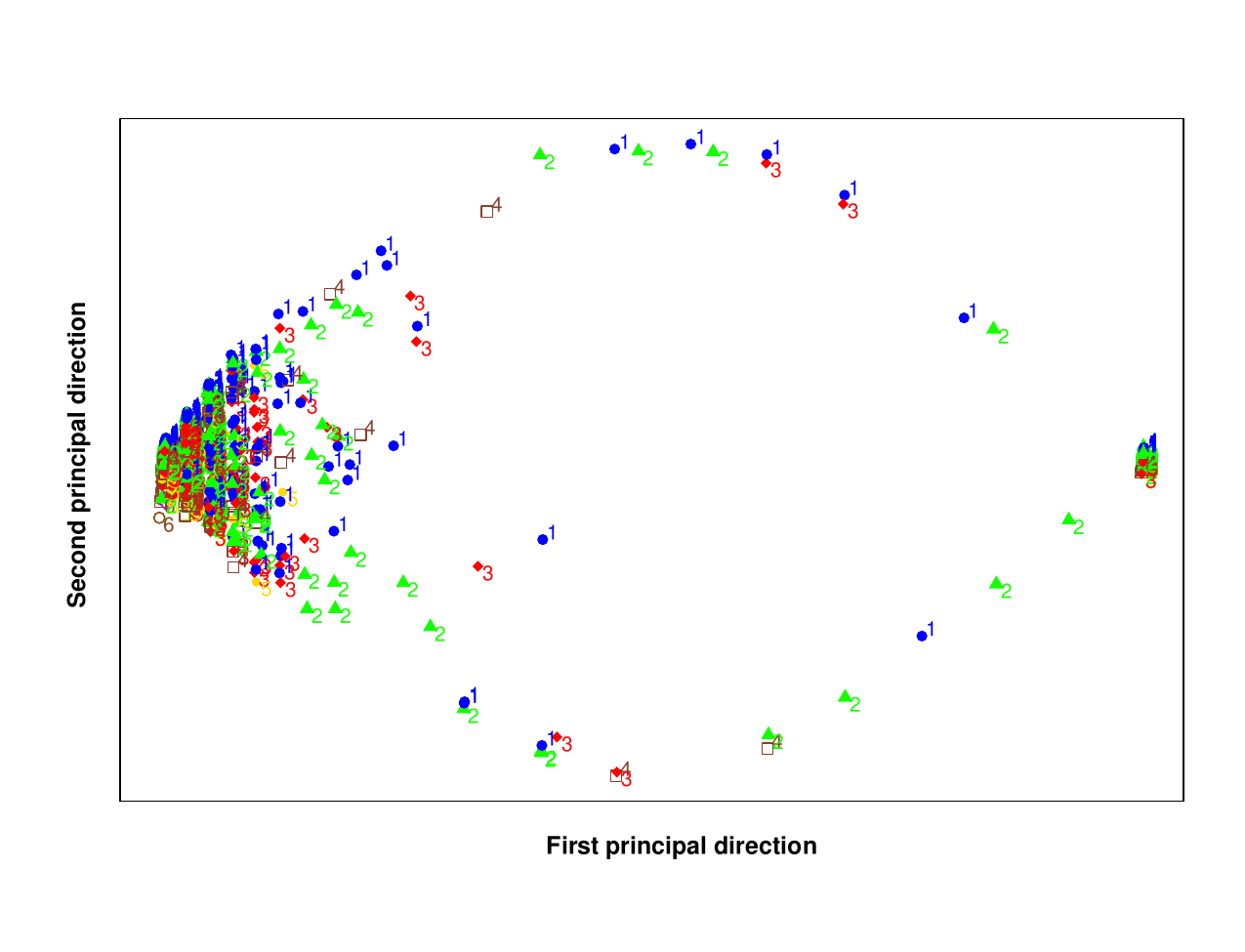}
\end{center}
\abovecaptionskip=-1pt
\caption{\it Principal component visualization of data. The points are drawn by their value along the top 2 principal directions, and labeled 
by Kp indices (rounded for clarity) and marked by different colors and shapes.} 
\label{figure:megPCA}
\end{figure}
\\\\
Figure~\ref{figure:megPCA} is a principal component visualization of the data. 
Together the top two principal directions captures 73.62\% of the variation in the data 
so the visualization should be a fairly faithful representation of the original data. 
The figure shows that the data (or the underlying pattern) is highly nonlinear and complicated: data points
with same Kp indices are scattering around while those with different Kp indices are highly mixed. This motivates us to use RF regression
in this work, as it can capture nonlinear patterns in the data. 
As our task is to build a model to separate data points corresponding to different Kp indices, the figure also shows that this task 
is challenging.
\section{Experiments}
\label{section:experiments}
\begin{figure}[!htb]
\centering
\begin{center}
\includegraphics[scale=0.65,clip, angle=0]{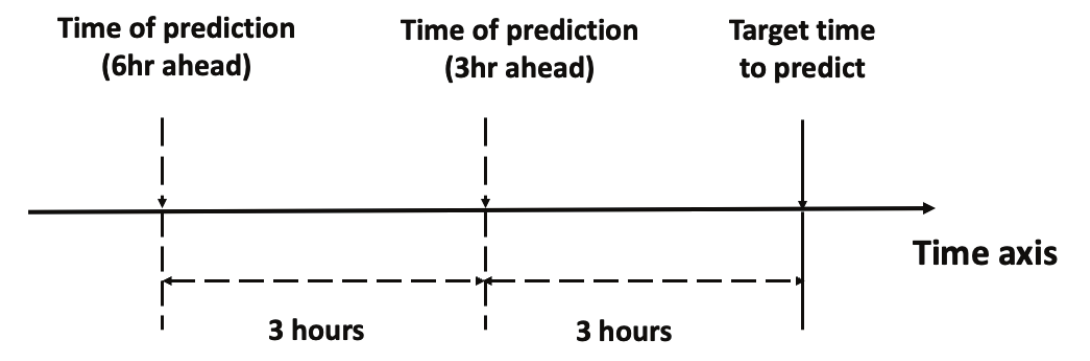}
\end{center}
\caption{\it Illustration of the difficulty in early prediction. For a given target time to predict, a 3-hour early prediction will use
all features up to 3-hour ahead and a 6-hour early prediction will use all features up to 6-hour ahead, and the difference is
the features associated with time between 3-6 hours ahead. } 
\label{figure:pastFeatures}
\end{figure}

\begin{figure}[!htb]
\centering
\begin{center}
\includegraphics[scale=0.63,clip, angle=0]{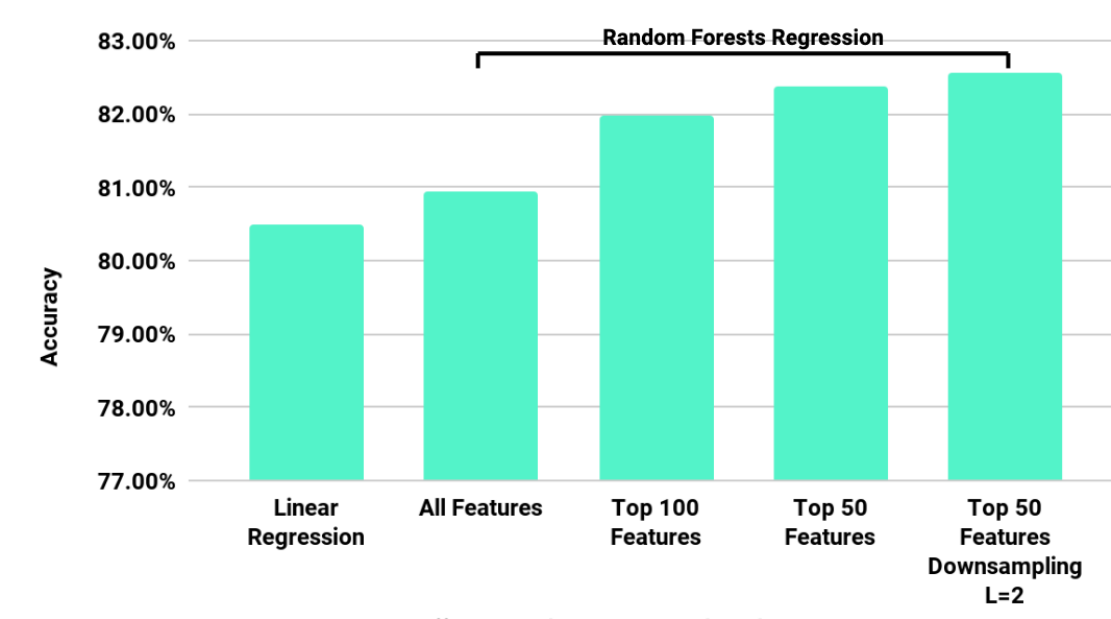}
\end{center}
\caption{\it Accuracy by different algorithms in predicting Kp index 3 hours in advance. When L=2, only 1/2 
of data with lower Kp indices was used in RF regression training (downsampling).} 
\label{figure:resultsComp}
\end{figure}

All the data used in our experiments are from 2021, and the first 9 months data are used for training and 
the next three months data for test. The number of trees in RF is fixed at 100 for all experiments, and the number of variables to select
at each node split is the default $p/3$ where $p$ is the number of features in the data. In this section, we report experimental results. 
\vspace{-15pt}
\begin{figure}[!htb]
\centering
\begin{center}
\includegraphics[scale=0.7,clip, angle=-90]{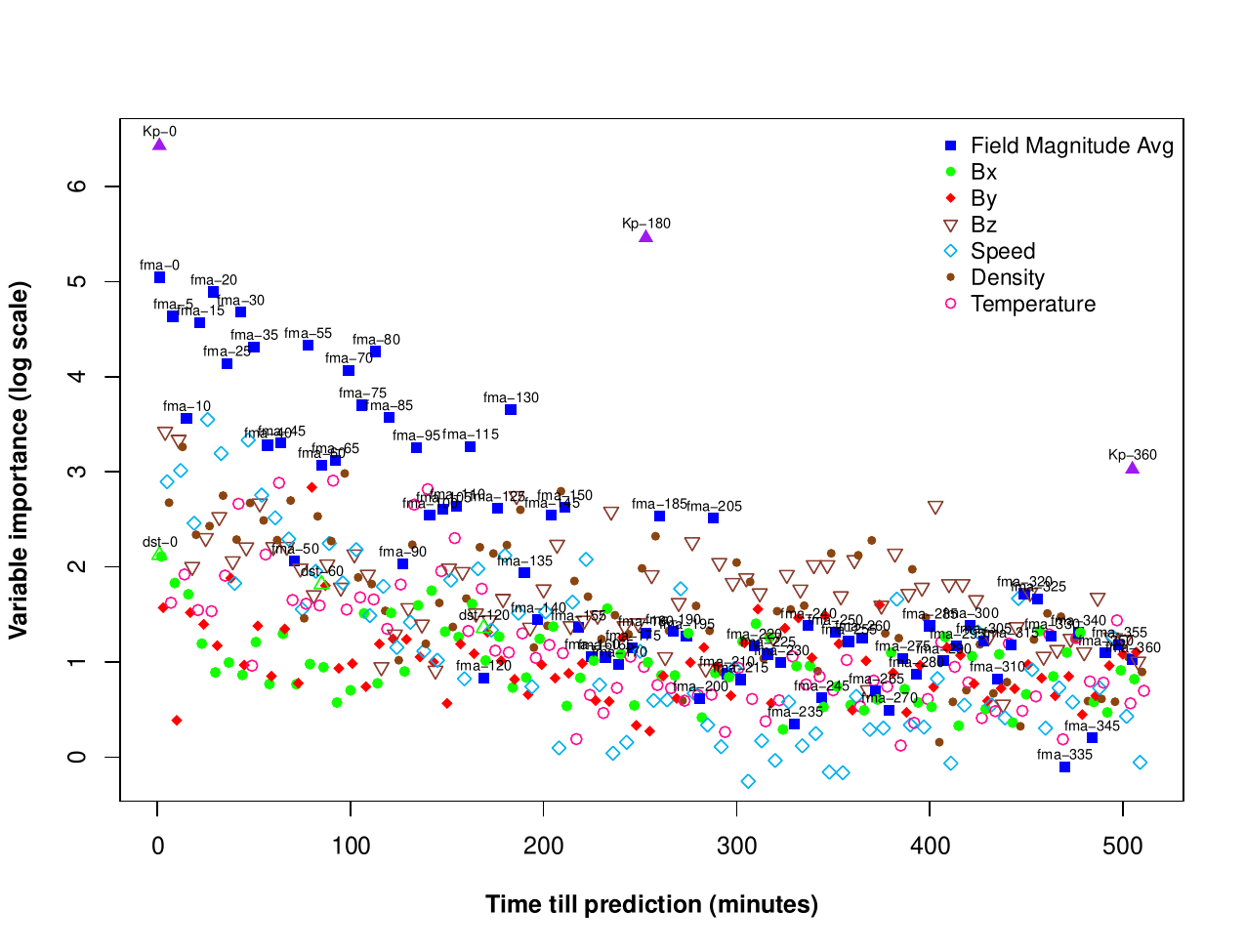}
\end{center}
\caption{\it Variable importance. Kp indices, FMAs, and other solar activity measurements taken at different time are arranged by the time 
they were measured, with more recent ones displayed first. The numbers in the label of points indicate the number of minutes till the current moment
at which time we are to predicate the geomagnetic disturbance 3 hours later. For clarity, features other than FMAs or dst's are not labeled.} 
\label{figure:varImp}
\end{figure}
\\\\
To select important features, we use the built-in RF feature importance tools to rank the features (which rank the features
by their contribution in the reduction of mean square errors in RF regression). We find that Kp indices before the time of prediction and field 
magnitude averages (FMA) are most important features. The importance of different features are shown in Figure~\ref{figure:varImp}.
We can see that variable importances decrease quickly with their associated time (till the current moment), meaning that
the further a variable is into the past, the less informative it would be for predictions of the future (e.g., magnitude of geomagnetic 
storms or Kp index 3 hours later). This increases the difficulty of earlier predictions, as the amount of historic data that can be used
is limited to the last few hours. 
Due to limited availability of Kp index measurements in current data, 3 hours in advance seems to have achieved the practical limit
in time for early prediction. 
Currently the Kp index is measured every 3 hours, an even earlier prediction (i.e., 6 hours ahead) with much less informative variables 
will lead to unsatisfactory results (not reported here due to limit in space). Compared to 3-hour early prediction, one can only use features 6-hour ahead with those corresponding to the last 3-6 hours will not be used, as illustrated in Figure~\ref{figure:pastFeatures}. However, as shown in Figure~\ref{figure:varImp}, features associated with the last 3-6 hours are much more important than those 6-hour earlier for a given target time to predict.
\\
\\
We assess the performance by the accuracy of prediction defined as the percentage of instances for which the predicted Kp index is within
1 of the actual Kp index. Figure~\ref{figure:resultsComp} shows the relative performance of different algorithmic choices. 
RF regression outperforms linear regression, which is consistent with the nonlinear pattern observed in Figure~\ref{figure:megPCA}. 
Feature selection of top 100 features improved the accuracy of RF regression; reducing to top 50 features further
improves the accuracy. This implies that the data,  in particular historic solar measurements in the very distant past, are less informative. Additional 
downsampling with top 50 features was able to achieve an accuracy of 82.55\% (we observe that upsampling of the minor classes can slightly improve 
accuracy, but results not reported here as not substantial). For the first time, all types of geomagnetic storms can be predicted
3-hour ahead, two hours earlier than previous work \cite{ZhuBillings2006}.
\\
\section{Conclusions}
\label{section:conclusions}
For the first time, all types of geomagnetic storms can be successfully predicted 3 hours ahead using machine learning algorithms. 
This was enabled by the use of big data from multiple sources from the world, coupled with techniques to "improve" the dataset for
sparse pattern detection, including feature selection and
downsampling. The algorithm achieves a prediction accuracy at 82.55\% by keeping only the most informative features (top 50
features), and downsampling data instances that occur more often (i.e., lower Kp indices) by half. 
\\
\\
The importance of variables from solar measurements decreases very quickly as the amount of advance in prediction time increases. Measurements taken 6 
hours in advance becomes barely informative, indicating that the achieved 3 hours in advance prediction has likely reached the practical limit, 
unless the Kp indices can be measured more frequently than the current every 3 hours.
\\
\\
Predicting geomagnetic storms accurately in advance will provide a sufficient warning time window for preparation.
This will greatly prevent or decrease damages to satellites, power grids, and communication infrastructures.
Our future work include the use of an ensemble model combining multiple predictive algorithms, or to provide a probabilistic confidence \cite{VovkShafer2008} for
specific predictions for example those for a very intense geomagnetic storm. Another possibility is to incorporate the physics 
of geomagnetic storms \cite{GonzalezJoselyn1994} in the prediction model to potentially improve accuracy or robustness of the prediction.
\section*{Acknowledgements}
I would like to thank my science research teacher, Mrs. Piscitelli, and my mentor Dr. Chen of Google Inc.
for their advice in project planning and suggestions in literature. I am also grateful to the editor, the associate editor,
and the anonymous reviewers for their constructive comments and suggestions.

\bibliographystyle{plain}

\end{document}